\documentclass[runningheads]{llncs}
\usepackage[T1]{fontenc}
\usepackage{graphicx}
\graphicspath{{images/}}
\usepackage[table]{xcolor}
\usepackage{nameref}
\usepackage{colortbl}
\usepackage{hhline}
\usepackage{float}
\begin{document}
%

\title{Exploring Scalability in Large-Scale Time Series in DeepVATS framework}

\titlerunning{Enhancing scalability in DeepVATS}
\titlerunning{Exploring Scalability in Large-Scale Timee Series in DeepVATS framework}
%
%
\author{Inmaculada Santamaria-Valenzuela\inst{1}\orcidID{0000-0002-7497-8795} \and
Victor Rodriguez-Fernandez\inst{1}\orcidID{0000-0002-8589-6621} \and
David Camacho\inst{1}\orcidID{0000-0002-5051-3475}}
\authorrunning{Inmaculada Santamaria-Valenzuela et al.}
%
\institute{ Dept. of Computer Systems, Universidad Politécnica de Madrid\\
\email{\{mi.santamaria,victor.rfernandez,david.camacho\}@upm.es}}
\maketitle              
\begin{abstract} 
Visual analytics is essential for studying large time series due to its ability to reveal trends, anomalies, and insights. DeepVATS is a tool that merges Deep Learning (Deep) with Visual Analytics (VA) for the analysis of large time series data (TS). It has three interconnected modules. The Deep Learning module, developed in R, manages the load of datasets and Deep Learning models from and to the Storage module. This module also supports models training and the acquisition of the embeddings from the latent space of the trained model. The Storage module operates using the Weights and Biases system. Subsequently, these embeddings can be analyzed in the Visual Analytics module. This module, based on an R Shiny application, allows the adjustment of the parameters related to the projection and clustering of the embeddings space. Once these parameters are set, interactive plots representing both the embeddings, and the time series are shown. This paper introduces the tool and examines its scalability through log analytics. The execution time evolution is examined while the length of the time series is varied. This is achieved by resampling a large data series into smaller subsets and logging the main execution and rendering times for later analysis of scalability. 
\end{abstract}

\section{Introduction}

\begin{figure}
    \includegraphics[width=1\linewidth]{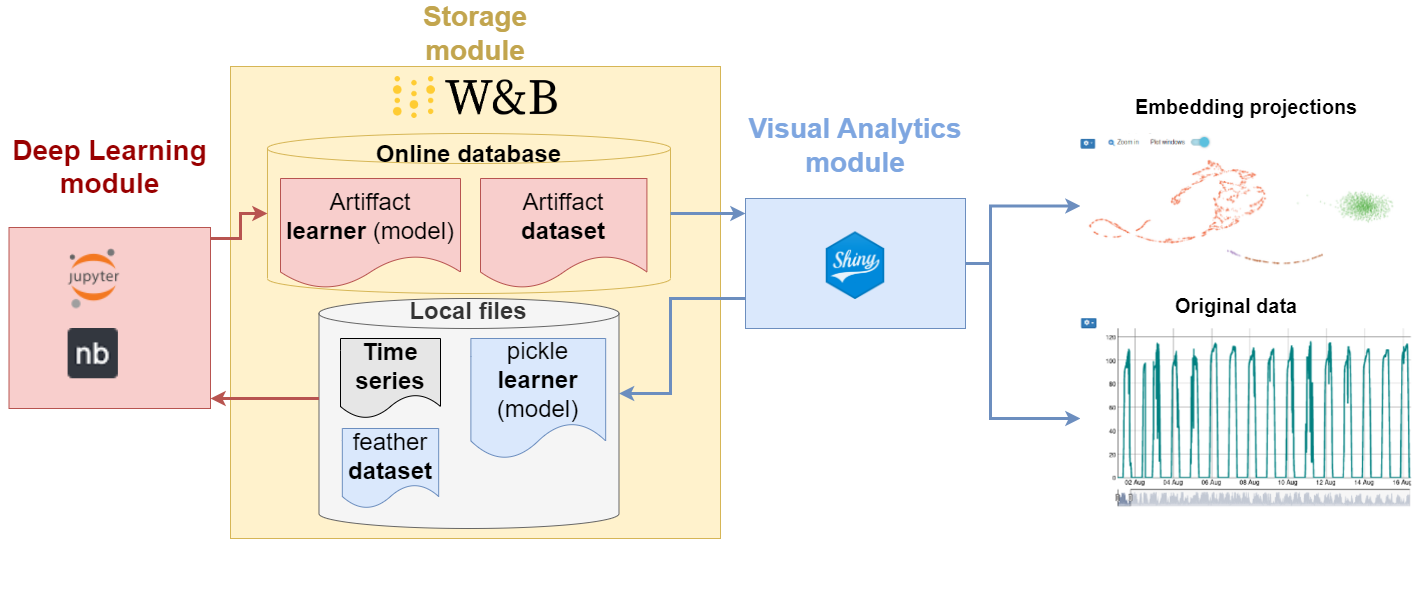}
    \caption{DeepVATS' global workflow. \label{fig:deepvats:schema:global}}
\end{figure}

Visual analytics in Deep Learning refers to the use of interactive visual interfaces and plots to identify emerging patterns and trends within the data, supporting its understanding. Dimensional reduction techniques are essential for the analysis and interaction with large time series data~\cite{ali2019timecluster}.

DeepVATS (Deep Visual Analytics for Time Series)~\cite{rodriguez2023deepvats} provides a tool inspired by TimeCluster~\cite{ali2019timecluster} that integrates Visual Analytics (VA) and Deep Learning (DL)  to extract valuable insights from Time Series (TS) through interactive projections and TS data plots. It uses dimensionality reduction (DR) techniques for efficient analysis and interaction with large time series. As Fig.~\ref{fig:deepvats:schema:global} shows, DeepVATS has three modules: DL, Storage and VA. These modules interact to perform three main tasks. These tasks include training neural networks to obtain meaningful data representations (embeddings); projecting the content of the trained neural network's latent space in two dimensions (2D) to detect clusters and anomalies; and finally, providing interactive plots to explore different perspectives of the projected embeddings.

The main goal of DeepVATS is to analyze large TS datasets. This article presents a scalability analysis performed using the Monash benchmark \cite{godahewa2021monash} to analyse future improvements that can be carried out. As ``Solar Power dataset (4 Seconds Observations)'' ~\cite{godahewa_2021_4656027} is the dataset that contains the largest TS in the benchmark, it has been selected as the case for use for the study.  This TS dataset represents the solar power production in MW recorded every 4 seconds over a year ($7397222\approx7.4$M elements in total). To get representative subsets for scalability analysis, this TS  has been resampled into datasets from 4 seconds to 10 minutes frequencies. This analysis has shown some performance problems. First, there is an issue of stability related to the use of cuml UMAP implementation~\cite{rapids_cuml_api_2024} that is unstable due to an internal issue ~\cite{cuml_issue_5474}. Two alternatives are proposed for future analysis. Second, the use of \texttt{reticulate} may add some time because of the communication between R and python. Thus, some operations have been joined or moved to be used from a pickled file instead of R variables to get better performance. Finally, some extra execution steps have been detected due to Shiny's reactiveness. This can be easily managed by a better use of the cache in the R shiny up, thus some alternatives are proposed within the analysis.

The remainder of the paper is organized as follows. Section \ref{sec:description}-\nameref{sec:description}: provides a description of DeepVATS including a step-by-step case of use. Section \ref{sec:scalability}-\nameref{sec:scalability}: examines the most representative steps within the execution. It includes the load of the dataset, obtainment of the embeddings, the calculus of the projections, the clustering, and the interactions with the interactive plots. Finally, section \ref{sec:conclusions}-\nameref{sec:conclusions} summarize the conclusions of the scalability analysis and future lines and expectations on the development of the tool.

\section{DeepVATS description \label{sec:description}}

\begin{figure}
    \centering
    \includegraphics[width=0.8\linewidth]{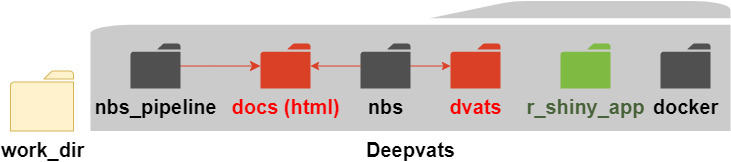}
    \caption{DeepVATS main folders schema and main jupyter notebooks for the executions.}
    \label{fig:deepvats:folders}
\end{figure}

The DeepVATS code base\footnote{The source code and the dockers used for the execution of the tool are hosted in the GitHub repository \url{https://github.com/vrodriguezf/deepvats}.} integrates three main modules: Deep Learning (DL) Module, Storage Module and Visual Analytics (VA) Module (See Fig.~\ref{fig:deepvats:schema:global}). 
The DL Module is a Python library implemented using Jupyter Notebooks ~\cite{jupyter_2024} and nbdev~\cite{nbdev_2024}. The Jupyter notebooks contained in the folder \texttt{nbs} are used for the generation of \texttt{dvats} library. This library is the base for both the \texttt{nbs\_pipeline} notebooks and the \texttt{r\_shiny\_app} (see Fig.~\ref{fig:deepvats:folders}). As detailed in the github repository, the docker containers must be preconfigured, assuming that another folder \texttt{work\_dir} already exists in the same path as the global ``Deepvats'' for the use of the tool functionalities.

\subsection{Deep Learning (DL) Module}
\begin{figure}
    \centering
    \includegraphics[width=0.5\linewidth]{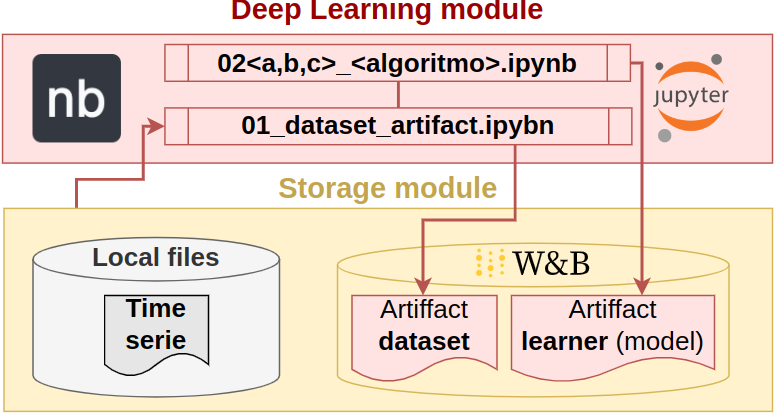}
    \caption{DeepVATS' load dataset workflow.}
    \label{fig:deepvats:schema:load_dataset}
\end{figure}

The DL module gives tools for loading datasets from/to the storage module and use them into the \texttt{nbs\_pipeline} folder notebooks. As Fig.~\ref{fig:deepvats:schema:load_dataset} shows, the first step is to execute \texttt{01\_dataset\_artifact}. This notebook assumes preconfigured dataset options defined on the \texttt{config/*.yml} files. Some examples with free accesible datasets are included via direct python execution. This execution will load a dataset from local to Weight \& Biases (\url{wandb.ai}) as an artifact. It can be  originally a ‘csv’, ‘txt’ or ‘tsf’ file. Now, we can use the \texttt{02\_<a,b>\_encoder\_<DCAE, MVP>} notebooks to train the associated fastai learners' models~\cite{fastai_2024}.  There are currently implemented two algorithms for data encoding: Deep Convolutional auto-encoder (DCAE) ~\cite{ali2019timecluster} and Masked Value Prediction ~\cite{rodriguez2023deepvats}. This three jupyter notebooks includes its equivalent using Sliding Window View for better performance. 
Once we got the model trained and loaded to W\&B, the visual application should be used. However, the functionalities can be tested in \texttt{03a\_embeddings}, which is based on TSAI’s \texttt{get\_acts\_and\_grads()} for extracting the embeddings from the trained model~\cite{tsai} and \texttt{04a\_dimensionality} \texttt{\_reduction} can be used for getting the embeddings from the learner and applying UMAP ~\cite{mcinnes2018umap}, TSNE or PCA (rapidsai’s cuml implementations ~\cite{rapids_cuml_api_2024}). Finally, \texttt{hdbscan} is used for clustering ~\cite{hdbscan_2024} and the plots are shown.  

In order to get dynamic plots depending on the dataset, the encoder, projection algorithms parameters and clustering parameters, the VA module is implemented. It includes their parameters so that different configurations can be tested and time series can be easily analyzed. This module is based on R and uses shiny for building up an app where a projection plot (with the projection of the dimensionally reduced embeddings) and a time series plots are shown. As it will be described in the next sections, both plots are interactive, so insightful details can be obtained from them. 

\subsection{Visual Analytics (VA) module}
\begin{figure}
    \centering
    \includegraphics[width=1\linewidth]{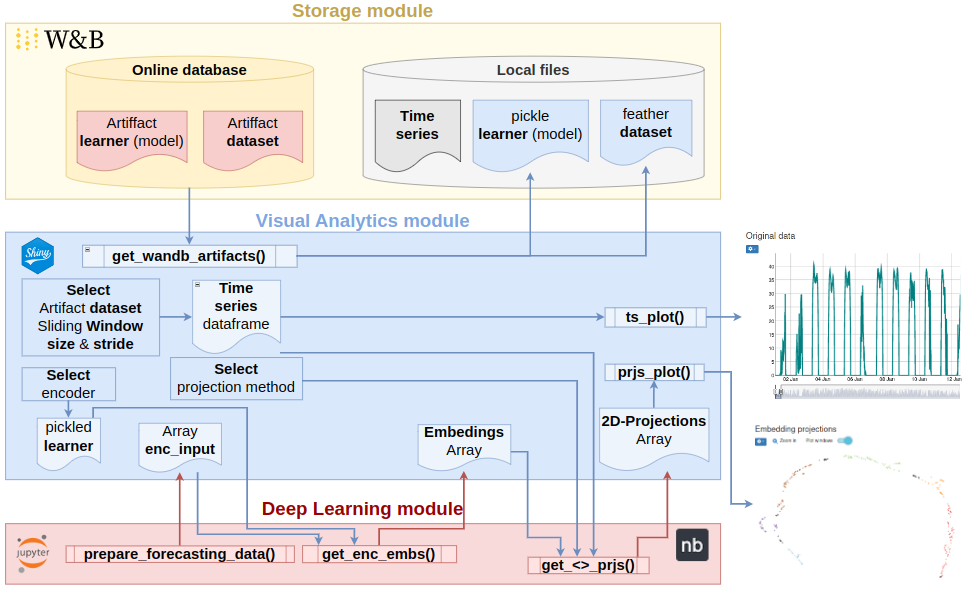}
    \caption{DeepVATS Visual Analytics (VA) module workflow and interaction with Storage module schema}
    \label{fig:deepvats:visual_analytics_module}
\end{figure}

\begin{figure}
    \centering
    \includegraphics[width=0.5875\linewidth]{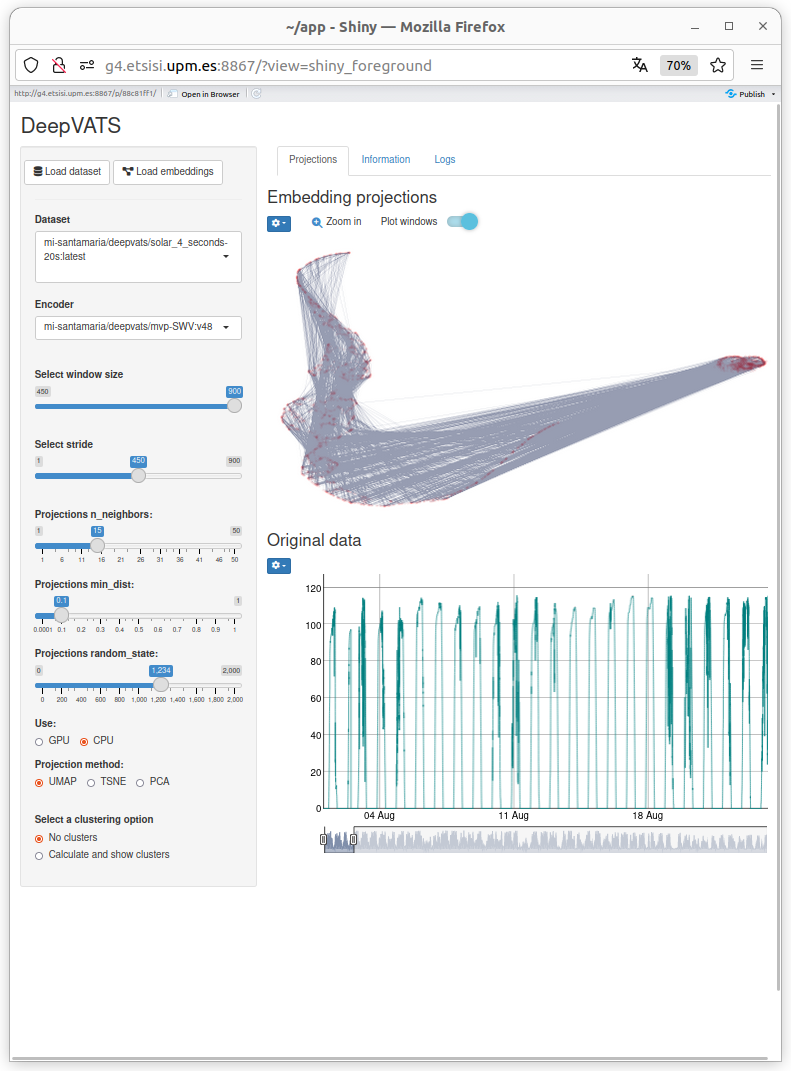}
    \caption{DeepVATS' Visual Analytics (VA) Module first view example.}
    \label{fig:deepvats:r_shiny_app:first_view}
\end{figure}
This section presents the VA module through images. As Fig. \ref{fig:deepvats:visual_analytics_module} show, the visual app does the following main steps. First, it downloads all artifacts from the W\&Biases system. Second, from a selected dataset and an associated encoder, it obtains the embeddings and projects them into a projection plot (PP) using \texttt{dvats}. At the same time, it generates the TS plot associated to the dataset artifact. The application also includes the option of clustering the projections for better analysis. To open the app we must run the r-studio-server docker and run \texttt{shiny::runApp(``app'')}. This will open a new window (permissions may be requested by the browser). 

This window contains the following main elements (see Fig.~\ref{fig:deepvats:r_shiny_app:first_view}: 
On the left, selectors for getting specific dataset, encoders and associated configurations: 
\begin{itemize}
    \item Encoder selector. It gives a list with all W\&B encoder artifacts trained for the loaded dataset. Default: last uploaded artifact for the dataset.
    \item Slider window size selector. Default: used for \texttt{02b\ prepare\_forecasting} \texttt{\_data()}. Maximum and minimum are proposed in the artifact configuration file.
    \item DR parameters slider selectors: \texttt{n\_neighbors}, \texttt{min\_dist} and \texttt{random\_state}.  Default: configured in the artifact configuration file.
    \item Projections algorithm selector: UMAP, TSNE, PCA
    \item GPU/CPU selector. Default: GPU.
    \item Clustering option: Calculate and show clusters. For clustering the embedding projections dataset.  When calculate and show cluster is selected, hdbscan parameters sliders are shown.
\end{itemize}

On the right side, there are three tabs. First, \texttt{projections} tab, which shows the interactive plots. Second, the \texttt{information} tab, which shows the metadata of the selected artifacts. Finally, the \texttt{Logs} tab, which is under development and summarises the main logs including the execution times. 

\begin{figure}
    \centering
    \includegraphics[width=0.23\linewidth]{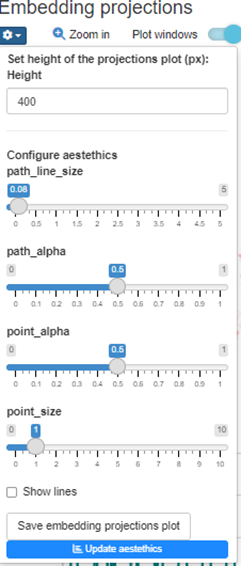}
    \includegraphics[width=0.41\linewidth]{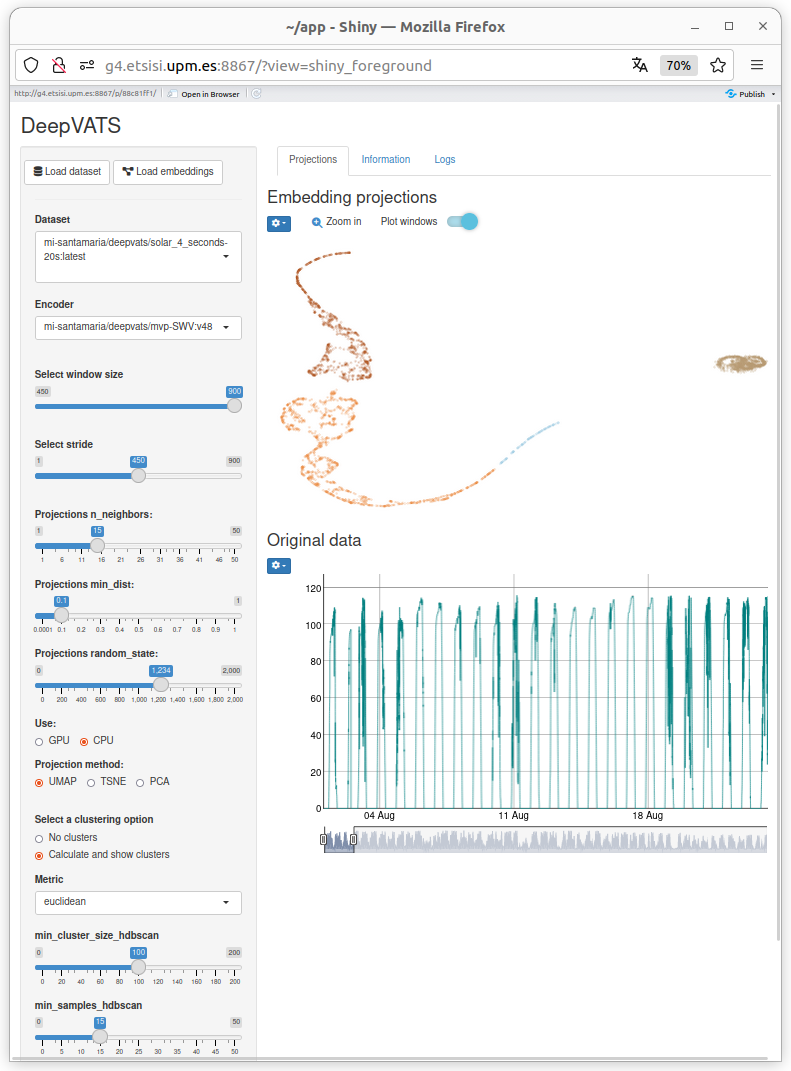}
    \caption{DeepVATS' Visual Analytics (VA) Module example of projections plot aesthetics configuration}
    \label{fig:deepvats:r_shiny_app:pp:aesthetics_configuration}
\end{figure}

Focusing on the \texttt{Projections} tab, we can interact with the plots in different manners. First, we can change the aesthetics through the embedding projections options menu for better visualisation (see Fig.~\ref{fig:deepvats:r_shiny_app:pp:aesthetics_configuration}). This includes different parameters for the aesthetics and a download button for saving the embedding projections plot. 

\begin{figure}
    \centering
    \includegraphics[width=0.25\linewidth]{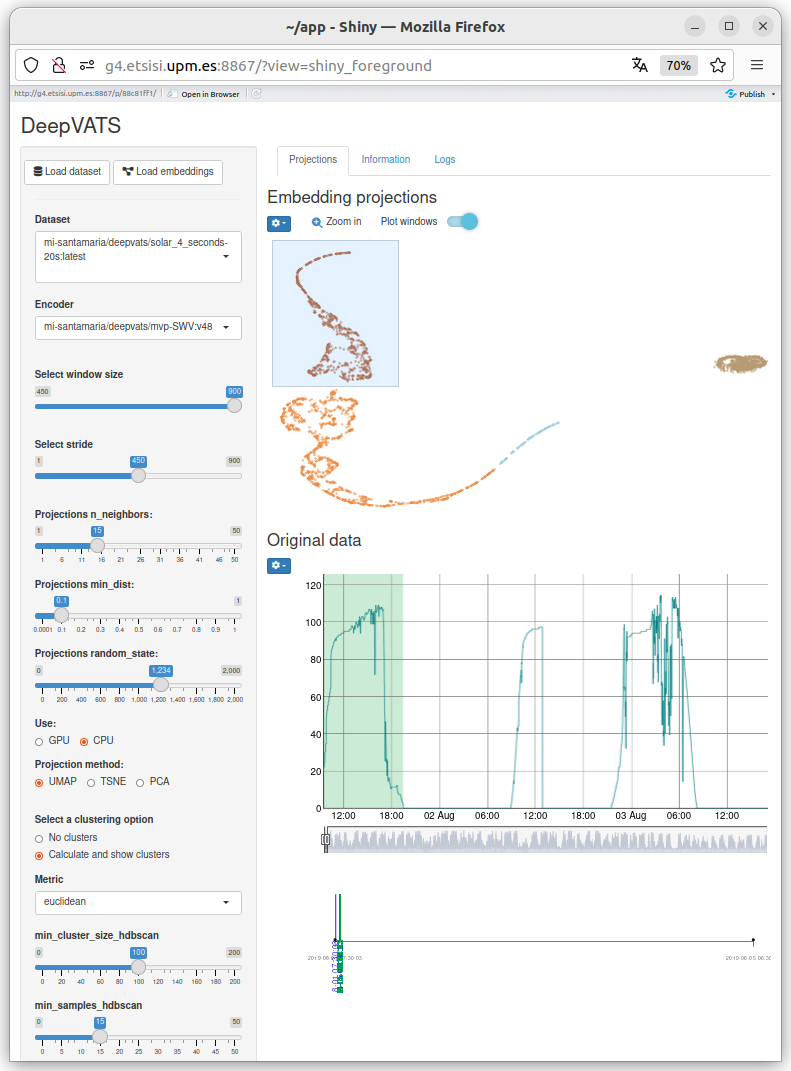}
    \includegraphics[width=0.7\linewidth]{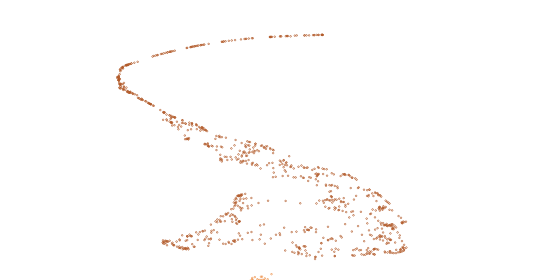}
    \caption{DeepVATS' Visual Analytics (VA) Module projections plot zoom in example}
    \label{fig:zoomin}
\end{figure}

Once we have the embeddings projections configured, we can get more insights on the relation between the embeddings and the TS data. By selecting points in the embeddings projections, their corresponding windows are displayed in the time series and viceversa. To avoid the application from crashing, the number of timepoints to display has been reduced to 10k. This also gives better rendering performance. The under-development lines below approximately show the windows in the same position as in the graphic timeline. The time selector can be used so that different windows can be analyzed. Also, zoom can be used for better comprehension of the embeddings structure by pushing the zoom in-zoom out button (see Fig.~\ref{fig:zoomin}).

The information tab shows the dataset and the encoder metadata and the Logs tab, which is under development, saves a datatable with basics logs of the app. 
\section{Scalability Analysis \label{sec:scalability}}

\begin{table}
\centering
\caption{Resamples of \texttt{Solar Power Dataset (4 Seconds Observations)} used for the scalability analysis. Seconds is the frequency generated by using the proposed frequency factor. The last column shows the number of elements of each dataset.}\label{tab1}
\begin{tabular}{|l|l|l|}
\hline
Seconds &  Frequency factor & Number of elements \\
\hline
4 & 1 & 7,397,222 \\
20 &  5 & 3,698,611 \\
60 & 15 & 493,149 \\
300 (5m) & 75 & 98,630 \\
600 (10m) & 150 & 49,315 \\
\hline
\end{tabular}
\end{table}

The main goal of deepVATS is to visually analyse large time series. The Monash benchmark \cite{godahewa2021monash} presents more than 20 datasets with different number of series and lengths, thus providing for good data for testing the app's performance. To test the most challenging case, the dataset with the longest time series has been used: Solar Power Dataset (4 Seconds Observations) \cite{godahewa_2021_4656027}. This dataset contains a single time series containing the solar power production in MW (megawatts) recorded with a frequency of 4 seconds along one year, obtaining more than 7M elements (see Table 1).  

To check the app's scalability for ensuring its performance in large time series, this dataset has been resampled into smaller datasets by selecting the dataset rows based in a frequency factor, defining the frequency factor ($f_{factor} = 150$) as the number we multiply frequency  by ($f = 4s$) so that desired frequency is obtained ($f = 4s \cdot 150 = 600s = 10m$). For simplicity, after checking that 10 minutes frequency had also been previously used in the Monash benchmark \cite{godahewa_2021_4656144}, factors $5$, $15$, $75$ and $150$ have been selected to obtain divisions of 4 seconds to 10 minutes frequencies. Table~\ref{tab1} shows the evolution in terms of number of elements. To get the execution times and logs, the following steps have been executed for each dataset: 

\begin{description}
    \item [Load the dataset from the feather file.] Select the desired dataset with the last associated encoder logged and check the time required to load the dataset from the binary file. The app will automatically get the embeddings from the encoder and apply UMAP using GPU for dimensionally reduction (DR) and will generate the associated projections and time series plots.

    \item [Change to PCA.] As will be detailed in the following sections, it is interesting to check the use of PCA and other DR algorithms.
    \item [Change to CPU.] It is interesting to check the app performance according to this selection.

    \item [Clustering.] To gain insight into the structure of the embeddings.

    \item [Selection of projection points.] The TS points associated to the selected points will be shaded in the TS plot. 
    \item [Plots interactions.]  To check the plots rendering after interactions, basic steps have been added: projections plot (PP) zoom in and zoom out, select a point on the time series plot so its associated projections points are remarked.
    \item [PP aesthetics updates.] To check rendering times for aesthetics updates, minimal changes have been selected: update point alpha aesthetic in PP and removing lines in PP for clearer visualisation.
    
\end{description}

The following sections analyses the different steps that the app executes when those steps are done: load dataset, get encoder embeddings, compute projections, compute and visualize plots and compute clusters.

\subsection{Load dataset}
\begin{figure}
    \centering
    \includegraphics[width=1\linewidth]{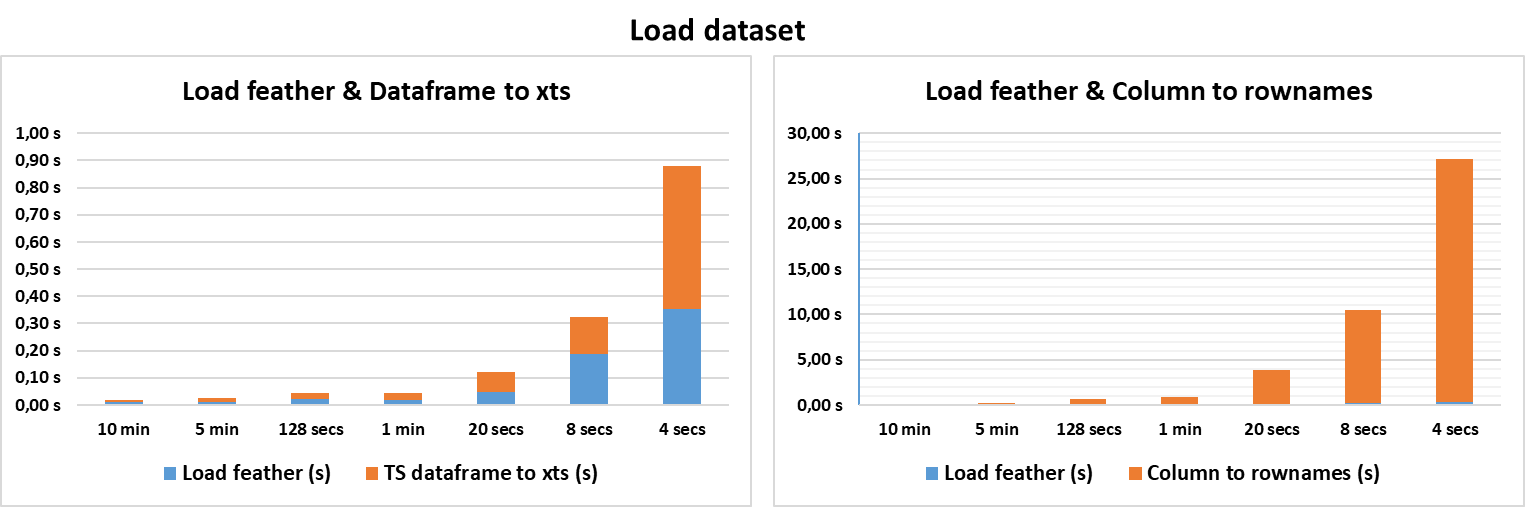}
    \caption{Load dataset aggregated time plot. In blue, read feather operation. In orange, on the left, time series dataframe conversion to xts. In orange, on the right, move the timeindex column to rownames operation.}
    \label{fig:scalability:load_dataset}
\end{figure}

TS plot is generated using \texttt{dygraph}~\cite{dygraphs_shiny}. This function expects to receive a dataframe with timestamps as rownames. Thus, a rownames conversion was initially added to move the time index column to rownames. This step rapidly increases the execution time when increasing the TS length. After taking insights on the execution progress, a dataframe to \texttt{xts} conversion was detected when calling dygraph. Fig.~\ref{fig:scalability:load_dataset} shows how explicitly making the conversion via \texttt{xts(TS\_dataframe,order.by=TS\_dataframe\$timeindex)} reduces the execution time up to $27$ seconds for the largest case, thus notably enhancing the app performance.

\subsection{Get embeddings}

\begin{figure}[H]
    \centering
    \includegraphics[width=0.45\linewidth]{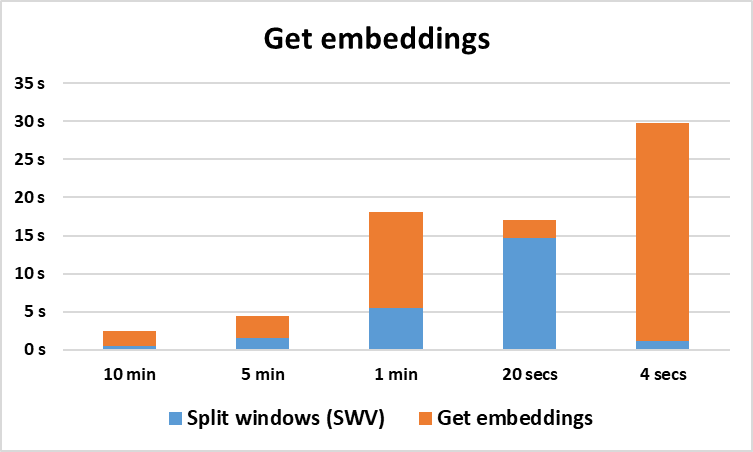}
    \caption{Aggregated plot for \texttt{get embeddings} step. In blue, the execution time consumed by the splitting window operation using Sliding Window View (in seconds). In orange, the execution time consumption for the get encoder embedings operation (in seconds).}
    \label{fig:scalability:get_embeddings}
\end{figure}

The time and memory consumption increases rapidly when the embeddings are obtained from the trained models. For memory handling and better R-python communication the implementation has been modified. Now, the the dataset input is stored as a \texttt{feather} file, enabling data loading and manipulation within python. Furthermore, the embeddings obtainment is chunked so that GPU memory errors are avoided. 

\subsection{Compute projections}
\begin{figure}
    \centering
    \includegraphics[width=0.325\linewidth]{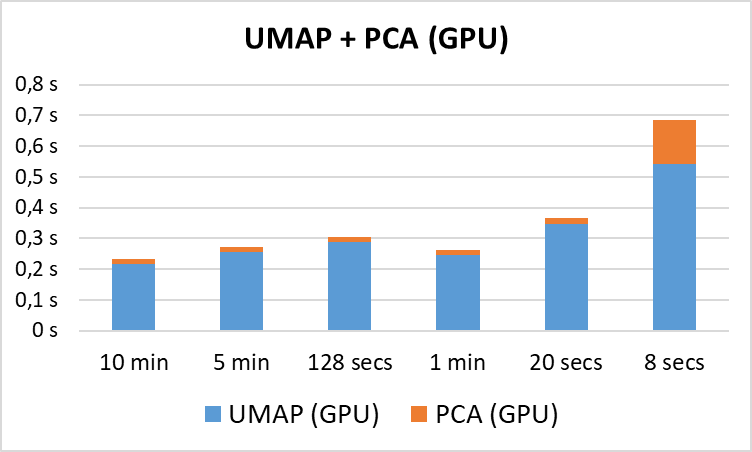}
    \includegraphics[width=0.325\linewidth]{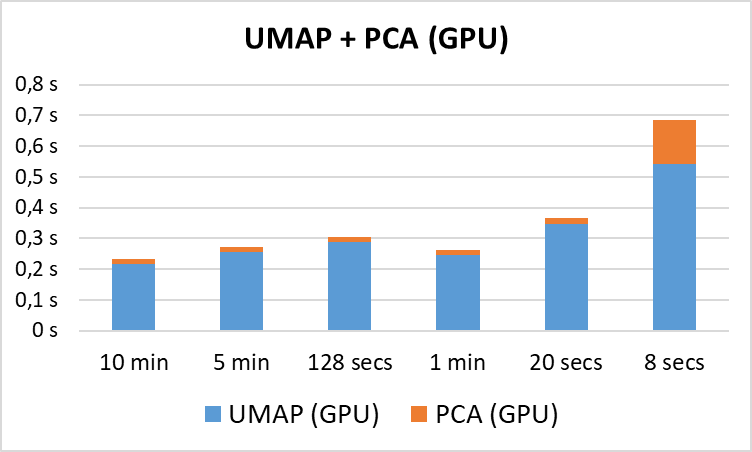}
    \includegraphics[width=0.325\linewidth]{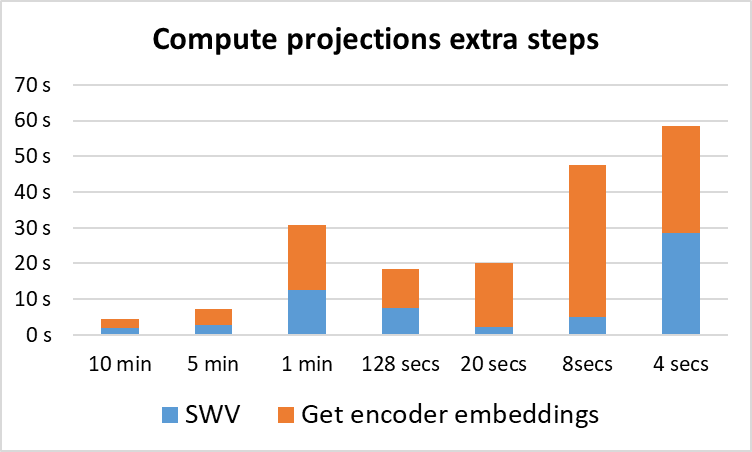}

    \caption{Compute projections. On the left UMAP + PCA (GPU) execution time aggregated plot. On the middle, the comparison with UMAP (CPU) executions. On the right, the execution time associated to extra executed steps.}
    \label{fig:scalability:compute_projections}
\end{figure}

The computation of projections is known to be much faster on a GPU than on a CPU. However, there exists a known bug in the GPU version that makes UMAP fail when using cuml UMAP (GPU) function that makes projections unstable and with low quality cluster distribution~\cite{raft_issue_313,cuml_issue_5474}. This quality can be checked in both the app logs and the \texttt{04} jupyter notebook, where \texttt{cluster\_score(prjs, clusters\_labels) = silhouette\_score(prjs, clusters\_labels)} is used to obtain a value in $[-1,1]$, with $-1$ being the worst score. 

To avoid this problem, the CPU version can be selected, but this results in an excessive execution time for a visual dynamic application. Thus, there exist two possible work lines. The first option is to change the GPU UMAP library that we use, for example, using Peter Eisenmann's parallel UMAP function~\cite{gpumap_peter}. The second option is to explore the solution proposed by Corey J. Nolet: Execute UMAP followed by PCA in order to obtain better results~\cite{cuml_issue_5474}. This has not been implemented yet, but, as Fig.~\ref{fig:scalability:compute_projections} shows, the time would still be much better with this option than using the CPU.

\subsection{Compute and visualise plots}

\begin{table}[H]
\centering
\caption{TS plot and PP plot rendering time after computing projections}\label{tab2}
\begin{tabular}{|l|l|l|l|l|l|}
\cline{2-6}
\multicolumn{1}{c|}{}& \multicolumn{5}{c|}{Time (seconds)} \\
\cline{2-6}
\multicolumn{1}{c|}{} & 10m & 5m & 1m & 20s & 4s \\
\hline

TSP rendering time & 4.11 & 5.25 & 20.48 &79.19 &  228.05 $\approx$ 4m\\
PP rendering time & 5.55 & 7.24 & 21.16 & 64.80 & 228.10 $\approx$ 4m\\
\hline
\end{tabular}

\vspace{2ex}
\caption{TS plot and PP plot computation time media and extra detected steps execution time in seconds}\label{tab3}
\begin{tabular}{
|l|
>{\columncolor{gray!25}\color{black}}l|
l|
>{\columncolor{gray!25}\color{black}}l|
l|
>{\columncolor{gray!25}\color{black}}l|
>{\columncolor{gray!25}\color{black}}l|
l|
>{\columncolor{gray!25}\color{black}}l|
l|
>{\columncolor{gray!25}\color{black}}l|}
\hhline{~----------}

\multicolumn{1}{c|}{} & \multicolumn{5}{c|}{\cellcolor{gray!25}\color{black}Time (seconds)} & \multicolumn{5}{c|}{Time (\% Relative to total)}\\
\hhline{~----------}
\multicolumn{1}{c|}{} & 10m & 5m & 1m & 20s & 4s & \%10m&\%5m&\%1m&\%20s&\%4s\\
\hline

TSP computation time & 0.03 & 0.03 & 0.08 & 5.45& 25.37 & 0.63&0.39&0.26&23.23&46.27\\

PP computation time & 2.16 & 3.28 & 12.46 & 0.93 & 0.05 &45.92&42.65&40.71&3.97&0.09\\
Extra steps & 2.51 & 4.38 & 17.06 & 29.4 & 29.83&53.45&56.95&59.03&72.78&53.63\\
\hline
\end{tabular}
\end{table}

A great reduction in the performance was detected when trying to load the Solar Power Dataset (4 Seconds Observations)  and interacting with the plots. This is reasonable as it is the largest one.   This is not a huge problem for small datasets, but causes the app to crash with large datasets. But our goal is to analyse these large datasets with the provided tool. Thus, this scalability analysis was started, and we detected that some reactive operations are being executed without need (see Table~\ref{tab3}). This adds a lot of time to the rendering of the plots. Also, a notable difference was noticed between execution and rendering time (see Table~\ref{tab2}), allowing nearly 4 minutes to render the plots for the 4 second frequency dataset. To avoid this problem, cache can be used in shiny to avoid plot recalculations. Also, the reduction of the number of showed points have reduced the renderization time. Therefore, the next step of the work is to analyse the use of more \texttt{reactiveVal}, and cache plots~\cite{shiny_global}, and to check more options for the reduction of the plotted points.

\subsection{Compute clusters}

The clusters computation is really fast even for large time series. However, some reactive functions have also been detected that affects to the performance of clustering step. Thus, the cache analysis is useful for enhancing this step too.
\section{Conclusions and future work \label{sec:conclusions}}

DeepVATS is a powerful tool designed for the easy visual analysis of both univariate and multivariate time series. It allows users to interact with the embeddings projection plot and the original time series data plot, facilitating the detection of patterns within the data. The scalability analysis has demonstrated an excellent performance for small datasets (ranging from 49.3K to 98.6K elements) and medium-sized datasets (up to 493.1K elements). However, performance issues appeared when processing larger datasets, with noticeable degradation at 3.7 million elements and application crashes at 7.4 million elements. 

To enhance the application's usability for large datasets, we propose the following development work lines. First, to eliminate redundant processes, the \texttt{reactive} variables should be checked to determine if they can be converted to \texttt{reactiveVal} to ensure the use of cached values. Also, \texttt{renderCachedPlot} is proposed as an alternative to \texttt{renderPlot} for the projections plot \cite{shiny_global}. Second, to ensure high-quality dimensionality reduction, two strategies are suggested: adopting an alternative UMAP implementation \cite{gpumap_peter}, and exploring the application of PCA followed by UMAP, rather than UMAP alone \cite{cuml_issue_5474}. 

These modifications are anticipated to give the application a substantial improvement in performance (e.g., the elimination of the additional 28 seconds required for compute projections for the 4 seconds frequency dataset). Also, greater stability in the GPU clustering is expected, making DeepVATS an more robust tool for the visual and interactive analysis of time series.

\subsubsection*{Acknowledgements.}
This work has been funded by Grant PLEC2021-007681 (XAI-DisInfodemics) and PID2020-117263GB-100 (FightDIS) funded by MCIN/ AEI/10.13039/501100011033 and by “ERDF A way of making Europe”, by the “European Union'' or by the “European Union NextGenerationEU/PRTR”, by Calouste Gulbenkian Foundation, under the project MuseAI - Detecting and matching suspicious claims with AI, and by "Convenio Plurianual with the Universidad Politécnica de Madrid in the actuation line of Programa de Excelencia para el Profesorado Universitario”.

%
%
%
%
%
%
\bibliographystyle{splncs04}
\bibliography{references}

\begin{thebibliography}{10}
\providecommand{\url}[1]{\texttt{#1}}
\providecommand{\urlprefix}{URL }
\providecommand{\doi}[1]{https://doi.org/#1}

\bibitem{ali2019timecluster}
Ali, M., Jones, M.W., Xie, X., Williams, M.: Timecluster: dimension reduction applied to temporal data for visual analytics. The Visual Computer  \textbf{35}(6-8),  1013--1026 (2019)

\bibitem{shiny_global}
Chang, W., Cheng, J., Allaire, J., Sievert, C., Schloerke, B., Xie, Y., Allen, J., McPherson, J., Dipert, A., Borges, B.: shiny: Web Application Framework for R (2022), \url{https://shiny.rstudio.com/}, r package version 1.7.4

\bibitem{hdbscan_2024}
scikit-learn contrib: scikit-learn-contrib/hdbscan. \url{https://github.com/scikit-learn-contrib/hdbscan} (2024), accessed: Jan. 04, 2024

\bibitem{gpumap_peter}
Eisennman, P.: p3732/gpumap. \url{https://github.com/p3732/gpumap} (2023), accessed: Jan. 04, 2024

\bibitem{nbdev_2024}
fast.ai: nbdev – create delightful software with jupyter notebooks. \url{https://nbdev.fast.ai/}, accessed: Jan. 04, 2024

\bibitem{fastai_2024}
fast.ai: fastai/fastai. \url{https://github.com/fastai/fastai} (2024), accessed: Jan. 04, 2024

\bibitem{godahewa_2021_4656027}
Godahewa, R., Bergmeir, C., Webb, G., Abolghasemi, M., Hyndman, R., Montero-Manso, P.: Solar power dataset (4 seconds observations) (Apr 2021). \doi{10.5281/zenodo.4656027}, \url{https://doi.org/10.5281/zenodo.4656027}

\bibitem{godahewa_2021_4656144}
Godahewa, R., Bergmeir, C., Webb, G., Hyndman, R., Montero-Manso, P.: Solar dataset (10 minutes observations) (Apr 2021). \doi{10.5281/zenodo.4656144}, \url{https://doi.org/10.5281/zenodo.4656144}

\bibitem{godahewa2021monash}
Godahewa, R., Bergmeir, C., Webb, G.I., Hyndman, R.J., Montero-Manso, P.: Monash time series forecasting archive. arXiv preprint arXiv:2105.06643  (2021)

\bibitem{jupyter_2024}
Jupyter, P.: Project jupyter. \url{https://jupyter.org}, accessed: Jan. 04, 2024

\bibitem{mcinnes2018umap}
McInnes, L., Healy, J., Melville, J.: Umap: Uniform manifold approximation and projection for dimension reduction. arXiv preprint arXiv:1802.03426  (2018)

\bibitem{tsai}
Oguiza, I.: tsai - a state-of-the-art deep learning library for time series and sequential data. Github (2023), \url{https://github.com/timeseriesAI/tsai}

\bibitem{raft_issue_313}
rapidsai: [bug] accuracy of lanczos solver. · issue \#313 · rapidsai/raft. \url{https://github.com/rapidsai/raft/issues/313}, accessed: Jan. 04, 2024

\bibitem{cuml_issue_5474}
rapidsai: [bug] different outputs for umap on cpu vs. gpu · issue \#5474 · rapidsai/cuml. \url{https://github.com/rapidsai/cuml/issues/5474}, accessed: Jan. 04, 2024

\bibitem{rapids_cuml_api_2024}
rapidsai: Rapids cuml api documentation. \url{https://docs.rapids.ai/api/cuml/stable/api/}, accessed: Jan. 04, 2024

\bibitem{rodriguez2023deepvats}
Rodriguez-Fernandez, V., Montalvo-Garcia, D., Piccialli, F., Nalepa, G.J., Camacho, D.: Deepvats: Deep visual analytics for time series. Knowledge-Based Systems  \textbf{277},  110793 (2023)

\bibitem{dygraphs_shiny}
Vanderkam, D., Allaire, J., Owen, J., Gromer, D., Thieurmel, B.: dygraphs: Interface to 'Dygraphs' Interactive Time Series Charting Library (2018), \url{https://github.com/rstudio/dygraphs}, r package version 1.1.1.6

\end{thebibliography}
\end{document}